\documentclass[twoside,twocolumn,10pt]{article}

\usepackage{wscg}           
\RequirePackage{ifpdf}
\ifpdf
 \RequirePackage[pdftex]{graphicx}
 \RequirePackage[pdftex]{color}
\else
 \RequirePackage[dvips,draft]{graphicx}
 \RequirePackage[dvips]{color}
\fi
\usepackage{amsmath}         

\usepackage[style=alphabetic]{biblatex}
\DeclareLabelalphaTemplate{
  \labelelement{
    \field[final]{shorthand}
    \field{label}
    \field[strwidth=3,strside=left,names=1]{labelname}
  }
  \labelelement{
    \field[strwidth=2,strside=right]{year}   
  }
}

\title{Similarity and symmetry measures based on fuzzy descriptors of image objects' composition}

\author{
\parbox{0.5\textwidth}{\centering
Marcin Iwanowski and Marcin Grzabka\\[1mm]
Institute of Control and Industrial Electronics, Warsaw University of Technology\\
ul.Koszykowa 75\\
00-662 Warszawa POLAND\\[1mm]
iwanowski@ee.pw.edu.pl
}}

\usepackage{url}
\makeatletter
\def\Uslash{\mathbin{\mathchar`\/}\@ifnextchar{/}{\kern-.15em}{}}
\g@addto@macro\UrlSpecials{\do \/ {\Uslash}}
\def\Ucolon{\mathbin{\mathchar`:}\@ifnextchar{/}{\kern-.1em}{}}
\g@addto@macro\UrlSpecials{\do : {\Ucolon}}
\makeatother

\begin{document}

\twocolumn[{\csname @twocolumnfalse\endcsname

\maketitle  

\begin{abstract}
\noindent
The paper describes a method for measuring the similarity and symmetry of an image annotated with bounding boxes indicating image objects. The latter representation became popular recently due to the rapid development of fast and efficient deep-learning-based object-detection methods. The proposed approach allows for comparing sets of bounding boxes to estimate the degree of similarity of their underlying images. It is based on the fuzzy approach that uses the fuzzy mutual position (FMP) matrix to describe spatial composition and relations between bounding boxes within an image. A method of computing the similarity of two images described by their FMP matrices is proposed and the algorithm of its computation. It outputs the single scalar value describing the degree of content-based image similarity. By modifying the method's parameters, instead of similarity, the reflectional symmetry of object composition may also be measured. The proposed approach allows for measuring differences in objects' composition of various intensities. It is also invariant to translation and scaling and -- in case of symmetry detection -- position and orientation of the symmetry axis. A couple of examples illustrate the method. 
\end{abstract}

\subsection*{Keywords}
image similarity, symmetry detection, object detection, fuzzy logic 
\vspace*{1.0\baselineskip}
}]

\section{Introduction}

In the paper, the method of content-based image similarity and symmetry measure is proposed\footnote{The authors conducted this research within the frames of the grant entitled "Powiedz mi, co widzisz" ("Tell me, what you see") of the POB SzIR of the Warsaw University of Technology.}. The subject of comparison is the composition of elements of the visual scene. It follows the way a human compares two images. For example, one may say that two images are close to another because on both, "there is a bicycle, a car, and a dog, the dog is below the bike, shifted to the left and a car is near the bike shifted above-left". 

The images under study have annotated data where the image pixel array is supplemented by the set of bounding boxes encasing the image objects. Such type of data is a typical result of, e.g., the object detection process. The method is focusing on the analysis of the mutual position of those objects. It applies fuzzy logic to describe the position of image objects' bounding boxes. The fuzzy estimation is based on the {\em fuzzy mutual position ($FMP$)} matrix that contains the fuzzy 2-D fuzzy position descriptors of every object pair~\cite{Iwanowski2021}. Such descriptor is computed for a given object in reference to all other ones. It consists of a fuzzy set that refers to spatial relation with other ones and the value of appropriate membership function. 

Two $FMP$ matrices, computed separately for all objects on both images, gather the information on the composition of two scenes. In the paper, a metric is proposed that considers the separately two aspects of the fuzzy relations between objects expressed by fuzzy 2-D descriptors, which is used to measure the similarity/symmetry of the composition of objects on both scenes. The first aspect refers to non-directional properties of the position, i.e., how far is one object from another, is inside or outside, and is called {\em locus descriptor}. The second aspect refers to the orientation and is called {\em orientation descriptor}. The comparison of images is performed by matching the composition of objects' pairs on both images and measuring the pairwise similarity using {\em matching matrices}. The accumulated pairwise similarity is finally normalized to get the output similarity of complete images. Thanks to the separation of the orientation aspect of the 2-D descriptor, it is possible to find the symmetric orientations for each type of 2-D descriptor. Consequently, one may evaluate the reflectional symmetry within the composition of image objects. 

The principal contribution of the research described in this paper is a method, allowing for applying the $FMP$ matrix to estimating the level of similarity and symmetries of the composition of visual scenes. The similarity and symmetry of images are considered here not -- as most of the known method does -- as direct similarity of the image content but in terms of the composition of objects detected on both images using, e.g., popular deep-learning-bases object detection algorithms. 

The paper is organized as follows. Section 2 contains the short survey of related papers. In section 3, the process of obtaining the $FMP$ matrix is recalled. Section 4 is the core one, including describing the proposed approach to comparing images and estimating their degree of symmetry. Section 5 is devoted to the presentation and discussion on example results. Finally, section 6 concludes the paper.   

\section{Related works}

Image comparison methods are used in many fields of image processing and analysis. They perform comparisons at various levels, starting from the lowest, where they are used to compute accumulated pixel-wise differences of images. At the highest level, image comparison methods are used for content-based image retrieval. More sophisticated image comparison methods work by building descriptors from image features at different levels and comparing them with various metrics and algorithms.  Low-level features may consist of color, texture, shape, or spatial location information. They can be extracted both locally or globally, at the image level. These features may be compared using distance in color space, histogram similarity measures, Minkowski, cosine, and Earth mover's distance distances, or non-parametric chi-square test~\cite{Rodden2000, Liu2006}. Some approaches like SSIM produce image similarity scores by comparing luminance, contrast, and structural features~\cite{Zhou2004,Wang2003}. At a higher level, region segmentation algorithms, like $k$-means, JSEG~\cite{Deng2001}, may utilize low-level features to extract and match similar areas in the images. 

Detecting symmetry in the image is a vital part of image understanding and can be used, e.g., for object detection and data compression. Symmetry can be divided into bilateral symmetry, rotational symmetry, and curve symmetry~\cite{Xiao2007}. In~\cite{Scognamillo2003} a two-stage algorithm has been proposed that locates visually salient features of the image by computing local energy function and then applies a Gaussian filter to extract the symmetry axis. An approach based on gradient orientation histogram and Fourier transform was proposed in~\cite{Sun1999}. The method described in~\cite{Loy2006} makes use of rotation invariant feature descriptors. The eigenvector decomposition of the covariance matrix to find bilateral symmetry hyperplanes was used in~\cite{OMara1996}.

Thanks to the rapid development of the deep-learning methods~\cite{LeCun2015}, various techniques for detecting and identifying the content of the digital image have been proposed. One of the most popular, primarily due to relatively high computational efficiency, is the object detection methods~\cite{Girshick2015,Redmon2016,Ren2017} that result in a handy set of bounding boxes encasing the image objects. The object detection methods are fast and -- in many cases -- able to process images in real-time. Their output may be further processed to get more sophisticated information on the higher-level properties of the visual scene. 

Describing the position of objects located within the image is another field of extensive study. Thanks to its ability to deal with the uncertainty that always accompanies a visual scene's interpretation, fuzzy logic has been often used as a principal tool in this field. In~\cite{Zhang1993,Miyajima1994} different fuzzy methods of scene representation, recognition, and description of an image with detected semantically consistent image regions. The latter paper introduced the histogram of orientations that was further investigated in many papers~\cite{Matsakis2001,Keller2000}. Detailed surveys of this field have been done in~\cite{Smeulders2000,Cohn2001,Bloch2005}.

\section{Extracting the fuzzy mutual positions matrix}

The notion of {\em fuzzy mutual position} (FMP) matrix that has recently been proposed in~\cite{Iwanowski2021} describe the content of the visual scene consisting of objects annotated by their bounding boxes. It is a data structure consisting of fuzzy membership function values for all pairs of image objects. Membership function values refer to fuzzy 2D position descriptors defined as fuzzy sets associated with particular spatial relations between two objects, e.g.," [one object] is located inside to the right [of the second one]". The details are this concept are recalled in following subsections. 

\subsection{Fuzzification of bounding box corners' coordinates}

Let focus at the beginning on the two-object case, let $B = \{ (x_B,y_B), (x'_B,y'_B )\}$ be a given bounding box defined by the position of its upper-left $(x_B,y_B)$ and lower-right $(x'_B,y'_B)$ corner. Let $R = \{ (x_R,y_R), (x'_R,y'_R )\}$ be the second bounding box -- the {\em reference} one. At first, the position of $B$ will be expressed relative to $R$ using the fuzzy position descriptor of its corner coordinates.

The relative position of any given image point $(\mathbf{x},\mathbf{y})$ in reference to $R$ is defined as:

\begin{equation}
    \mathbf{x} = 2 \cdot \frac{x - x_R}{x'_R - x_R} - 1 \; ; \;\mathbf{y} = 2 \cdot \frac{y - y_R}{y'_R - y_R} - 1.
    \label{eq:relative}
\end{equation}

Using the above equation, the coordinates of corners of the bounding box $B$ are transformed into relative ones:

\begin{equation}
    B_R = \{ (\mathbf{x}_{B \rightarrow R},\mathbf{y}_{B \rightarrow R}),  (\mathbf{x}'_{B \rightarrow R},\mathbf{y}'_{B \rightarrow R}) \},
    \label{eq:br}
\end{equation}

\noindent where $\mathbf{x}_{B \rightarrow R}$ stands for the relative coordinate of the given corner of the bounding box $B$ with bounding box $R$ used as the reference one.

The relative position $(\mathbf{x},\mathbf{y})$ is next fuzzyfied~\cite{Zadeh1965} separately for each axis into one of 5 fuzzy sets associated with the relative position of a point in relation to the reference bounding box $R$. They are described using trapezoidal membership functions. The following fuzzy sets are taken into account: {\em inside (i)}, {\em edge (e)}, {\em close (c)}, {\em near (n)}, and {\em far (f)}. The fuzzy values describe the distance zone counted starting from the centroid of $R$. In order to fully describe the position, in addition, the direction in which the point is located is coded using one of four values: {\em left (l)}, {\em right (r)}, {\em above (a)}, and {\em below (b)}. Fuzzification is performed for each axis separately, so finally there are two sets of fuzzy descriptors of corners of $B$, describing position along the x-axis: $D_{0x}  =  \{fl,nl,cl,el,il,ir,er,cr,nr,fr\}$ and along y-axis: $D_{0y}  =  \{fa,na,ca,ea,ia,ib,eb,cb,nb,fb\}$. As a result of the fuzzification process, each corner of the bounding box $B$ relative to $R$ along each of axes is described using one or two linguistic variables referring to point position descriptors associated with fuzzy sets and appropriate membership function values. These variables are {\em fuzzy descriptors of a point (corner)} relative to reference object $R$. 

\subsection{Determining the 2-D fuzzy bounding box position}

In the second step, starting from the corners' fuzzy descriptors, the next level descriptors are computed -- fuzzy {\em descriptors of edges} of $B$ relative to $R$. They are further referred to as {\em 1-D descriptors} as they reflect a bounding box's positions along single axes of the image coordinate system. Each bounding box is defined by four groups of corner descriptors (two for x-coordinates, two for y-coordinates, each of which consists of 1 or 2 descriptors), from which two groups of edge ones are derived. Depending on the position of an edge relative to the equivalent edge of the reference bounding box, the 1-D descriptor belongs to the following groups:

\begin{enumerate}
    \item {\bf outside position} -- edge of $B$ is located outside the edge of $R$ without any intersection. There are three such descriptors, depending on how far $B$ is from $R$: {\em far} ({\it FA}), {\em near} ({\it NE}), {\em close} ({\it CL}). In addition, the list of outside positions contain one additional case, when the edge of $R$ is included in the edge of $B$: descriptor {\em longer} ({\it LO})
    \item {\bf crossing position} -- edge of $B$ intersects the edge of $R$: possibly -- {\em touching} ({\it TO}) and actually -- {\em crossing} ({\it CR}), 
    \item {\bf inside position} -- edge of $B$ is entirely included in the edge of $R$: close to one of boundaries -- {\em inside} ({\it IN}) or around the center of $R$ -- {\em shorter}({\it SH}).
\end{enumerate}

The above descriptors are used along with orientation indicators that reflect the direction of a given property (one of 9 above listed). There are 4 indicators that follows the principal directions: {\em left} ({\it L})--{\em right} ({\it R}), for the x-axis (set of all 1-D descriptors along this axis will be denoted as $D_{1x}$) and {\em above} ({\it A})--{\em below} ({\it B}) for the y-axis ($D_{1y}$). There exist, however, some cases when exact direction cannot be determined. In these cases, the simplified indicators are used: {\em horizontal} ({\it H}) or {\em vertical} ({\it V}). They are accompanying descriptors {\it SH, SA, LO}.

The 1-D descriptors are computed based on fuzzyfied coordinates of bounding box corners -- upper left and lower right (eq.~\ref{eq:br}), which also depict the position of edges of the bounding box. They are ordered: $x_{B \rightarrow R} \leq x^{\prime}_{B \rightarrow R}$ and $y_{B \rightarrow R} \leq y^{\prime}_{B \rightarrow R}$, which implies that not all combinations of fuzzy descriptors of corners have to be considered (e.g. the left edge can only be located on the left side of the right edge -- cannot be located on the right side of the latter). Since the number of corner descriptors equal $|D_{0x}|=|D_{0y}|=10$, the total number of possible combinations equals $\sum_{i=1}^{10} i = 55$. Based on those combinations, the values of particular 1-D (edge) descriptors are computed using the {\em fuzzy associative matrix} ($\mathbf{FAM}$), which allows for finding the appropriate 1-D descriptor for each pair of edge descriptors. In this case, it is a 10x10 matrix with 55 relevant values.

Relations between two point descriptors of the horizontal direction (x-axis) in the bounding box $B$ ($\mathbf{x}_{B \rightarrow R}$ and $\mathbf{x}'_{B \rightarrow R}$) and 1-D position descriptor of the edge stretched between $\mathbf{x}_{B \rightarrow R}$ and $\mathbf{x}'_{B \rightarrow R}$ in relation to the horizontal edge of $R$ are shown partially in Table~\ref{tab:reasoning1}. It exhibits the fuzzy associative matrix $\mathbf{FAM}_1$, rows of which refer to the fuzzy position of $\mathbf{x}$, columns to the fuzzy position of $\mathbf{x}'$. Elements of this matrix contain the fuzzy 1-D descriptor of the edge. Similar matrix for vertical axis is obtained by replacing orientation indicators: {\it l} $\leftrightarrow$ {\it a}, {\it r} $\leftrightarrow$  {\it b}, {\it L} $\leftrightarrow$ {\it A}, {\it R} $\leftrightarrow$ {\it B}, {\it H} $\leftrightarrow$ {\it V}. 

\begin{table}[]
    \centering \tiny
    \begin{tabular}{|c||c|c|c|c|c|c|c|c|}\hline
           & {\it fl}  & {\it nl}  & {\it cl}  & {\it el}  & {\it il}  & {\it ir}  & {\it er}  & {\it  cr}   \\ \hline \hline
  {\it fl} & \it{FA/L} & \it{NE/L} & \it{CL/L} & \it{TO/L} & \it{CR/L} & \it{CR/L} & \it{CR/L} & \it{LO/H}  \\ \hline
  {\it nl} & ---       & \it{NE/L} & \it{CL/L} & \it{TO/L} & \it{CR/L} & \it{CR/L} & \it{CR/L} & \it{LO/H}  \\ \hline
  {\it cl} & ---       & ---       & \it{CL/L} & \it{TO/L} & \it{CR/L} & \it{CR/L} & \it{CR/L} & \it{LO/H}  \\ \hline   
  {\it el} & ---       & ---       & ---       & \it{TO/L} & \it{IN/L} & \it{IN/L} & \it{SA/H} & \it{CR/R}  \\ \hline
  {\it il} & ---       & ---       & ---       & ---       & \it{IN/L} & \it{SH/H} & \it{IN/R} & \it{CR/R}  \\ \hline 
  {\it ir} & ---       & ---       & ---       & ---       & ---       & \it{IN/R} & \it{IN/R} & \it{CR/R}  \\ \hline   
  {\it er} & ---       & ---       & ---       & ---       & ---       & ---       & \it{TO/R} & \it{TO/R}  \\ \hline  
    \end{tabular} \linebreak
    \caption{A part of $\mathbf{FAM}_1$ for estimating the 1-D fuzzy position descriptors, for horizontal axis (reduced due to space limitations, for complete 10x10 version see~\cite{Iwanowski2021}).}
    \label{tab:reasoning1}
\end{table}

The $\mathbf{FAM}_1$ is used to compute membership functions to fuzzy sets related to particular 1-D descriptors. This computation is performed using Mamdani fuzzy reasoning (min-max principle):

\begin{equation}
     \mu_{d}(\mathbf{p},\mathbf{p}')=\max \left\{\min(\mu_{d_1}\left(\mathbf{p}),\mu_{d_2}(\mathbf{p}'\right) \right\},
\end{equation}

\noindent where $\max\{\}$ is computed for all $d_1,d_2$ such that $\mathbf{FAM}_1(d_1,d_2) = d$. In the case of x-axis $\mathbf{p} \equiv \mathbf{x}_{B \rightarrow R}$,  $\mathbf{p}' \equiv \mathbf{x}'_{B \rightarrow R}$, $d_1,d_2 \in D_{0x}$, and $d \in D_{1x}$. In the case of y-axis $\mathbf{p} \equiv \mathbf{y}_{B \rightarrow R}$,  $\mathbf{p}' \equiv \mathbf{y}'_{B \rightarrow R}$, $d_1,d_2 \in D_{0y}$, and $d \in D_{1y}$. 

In the third and last stage of processing, the final 2-D set of descriptors of $B$ relative to $R$ is computed starting from two 1-D descriptors using another fuzzy associative matrix $\mathbf{FAM}_2$. The 2-D descriptor contains the information on the complete bounding-box 2-D position $B$ relative to $R$. 2-D descriptors express two types of spatial properties of the object $B$ relative to $R$: locus-based and orientation-based. {\em Locus-based} properties indicate where an object is located relative to the reference one and/or position without considering the direction. Some of them follow the concept of particular 1-D descriptors: {\em far} (FA), {\em near} (NE), {\em close} (CL), {\em touching} (TO), {\em crossing} (CR), {\em inside} (IN). The next two also follow the 1-D descriptors but do not express any directional properties: {\em same} (SA),  and {\em larger} (LG). The latter is equivalent to 1-D descriptor {\em longer} ({\it LO}), but its name fits better to the 2-D position. The last descriptor -- {\em split} (SP) refers to the case when the given object splits the reference one into two parts, which do not have a 1-D equivalent. 

\begin{figure}
 \centering
  \includegraphics[width=0.49\textwidth]{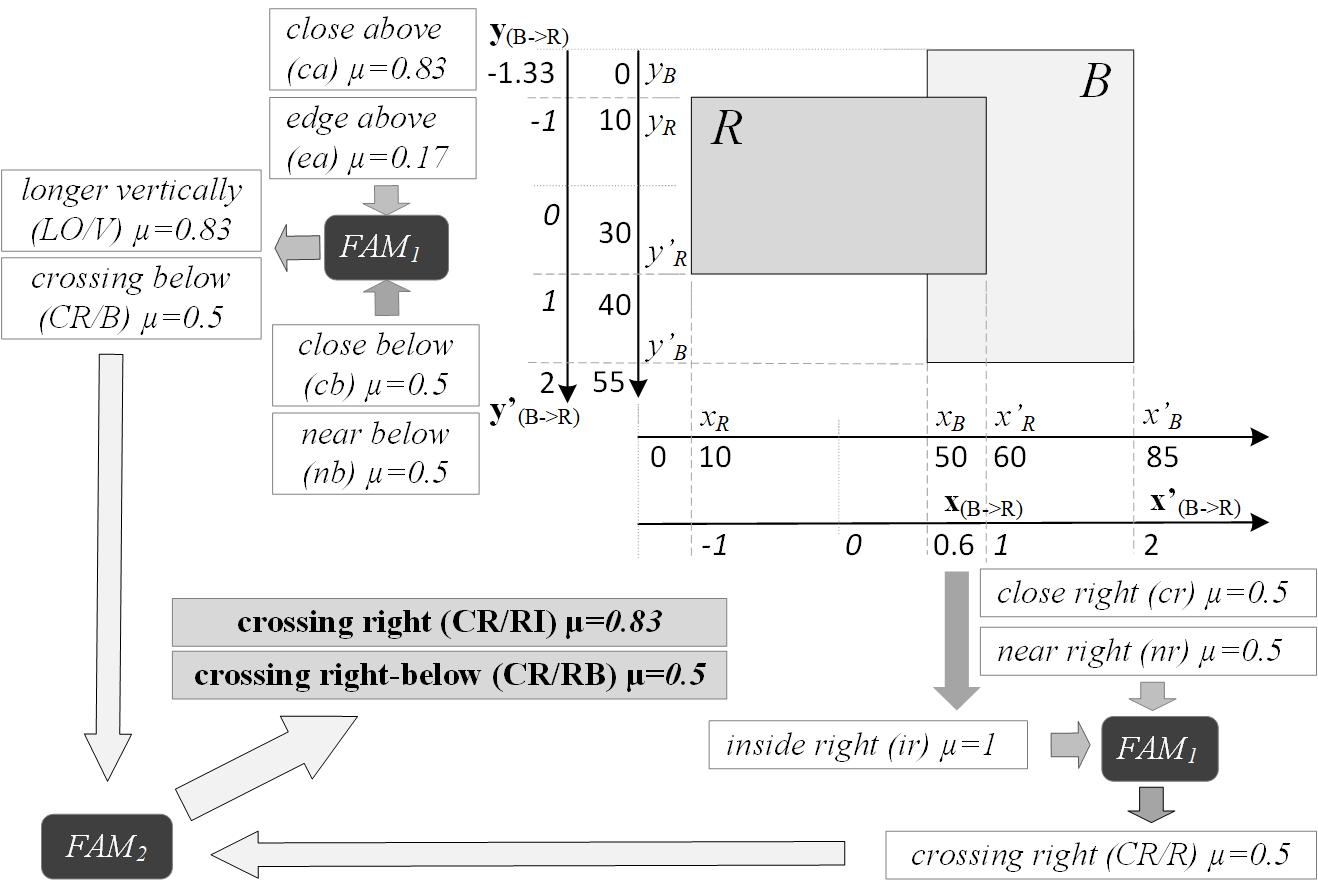} 
  \caption{Illustration of the process of generating the 2-D fuzzy descriptors of example bounding box $B$ relative to the reference $R$ consisting of tranformation of coordinates, fuzzification of bounding-boxes' corners, 1st reasoning resulting in estimated fuzzy 1-D position of edges and 2nd reasoning resulting in final 2-D descriptors.}
  \label{fig:reasoning3}
\end{figure}

The {\em orientation-based} descriptors indicate a direction in which the given locus-based descriptor property is fulfilled. There is a total number of 11 orientation-based descriptors. Eight of them indicate the primary directions:  {\em left} (LE), {\em left-above} (LA), {\em above} (AB), {\em right-above} (RA), {\em right} (RI), {\em right-below} (RB), {\em below} (BE) and {\em left-below} (LB) and are used together with locus-based descriptors: FA, NE, CL, TO, CR, and IN. Moreover, descriptors LE, RI, AB, and BE are used with the SP locus-descriptor to indicate splitting the reference object close to the particular edge. The first of the remaining three descriptors reflect the configuration when given bounding box have the same position of the centroid and same proportions: {\em centered} (CE) and is used together with locus-descriptors IN, SA, and LG. The last two direction descriptors {\em horizontal} (HO) and {\em vertical} (VE) are used only in combination with SP locus-descriptor to express the case of splitting of the reference bounding boxes in the middle of its either height or width. Considering all the above, the total number of 2-D descriptors -- combinations of locus-based and orientation-based -- equals $|D_2|=57$. Considering the fact that $|D_{1x}|=|D_{1y}|=15$, the size of $\mathbf{FAM}_2$ is 15x15 -- it contain all combinations of 1-D descriptors along x- and y-axis. A part of $\mathbf{FAM}_2$ matrix is shown in Table~\ref{tab:reasoning2}. 

Simmarily to the 1-D case, the 2-D descriptor is obtained using the Mamdani reasoning, this time using $\mathbf{FAM}_2$ matrix. For the bounding box $B$ defined by the Eq.~\ref{eq:br}, the membership function value of a 2-D descriptor $d$ is equal to:

\begin{equation}
     \mu_{d}(B_R)=\max \{\min(\mu_{d_x}(\mathbf{x}_{B \rightarrow R},\mathbf{x}'_{B \rightarrow R}),\mu_{d_y}(\mathbf{y}_{B \rightarrow R},\mathbf{y}'_{B \rightarrow R}) \},
     \label{eq:d2}
\end{equation}

\noindent where $\max\{\}$ is computed for all $d_x \in D_{1x}$ and $d_y \in D_{1y}$ such that $\mathbf{FAM}_2(d_x,d_y) = d$. The fuzzy relation $d$ belongs to the set of all 2-D fuzzy relations $D_2$.  The complete process of obtaining the 2-D fuzzy descriptors for an example object is shown in Fig.~\ref{fig:reasoning3}

With the usage of two $\mathbf{FAM}$ matrices, which are the loo-up-table ones, the calculations of fuzzy descriptors are fast. Nevertheless, one may speed it up even more by joining both matrices into a single 4-dimensional data structure, indexed by fuzzified corner coordinates (two along the x-axis and two along the y-axis). 

\begin{table}[]
\centering \tiny
\begin{tabular}{|c||c|c|c|c|c|c|c|}
\hline	
  &\it{CL/L} & \it{TO/L} & \it{CR/L} & \it{IN/L} & \it{SH/H} & \it{SA/H} & \it{LO/H} \\ \hline \hline
\it{FA/A} &  FA/LA &  FA/AB &  FA/AB &  FA/AB &  FA/AB &  FA/AB &  FA/AB \\ \hline
\it{NE/A} &  NE/LA &  NE/LA &  NE/AB &  NE/AB &  NE/AB &  NE/AB &  NE/AB \\ \hline
\it{CL/A} &  CL/LA &  CL/LA &  CL/AB &  CL/AB &  CL/AB &  CL/AB &  CL/AB \\ \hline
\it{TO/A} &  CL/LA &  TO/LA &  TO/LA &  TO/AB &  TO/AB &  TO/AB &  TO/AB \\ \hline
\it{CR/A} &  CL/LE &  TO/LA &  CR/LA &  CR/AB &  CR/AB &  CR/AB &  CR/AB \\ \hline
\it{IN/A} &  CL/LE &  TO/LE &  CR/LE &  IN/LA &  IN/AB &  IN/AB &  SP/AB \\ \hline
\it{SH/V} &  CL/LE &  TO/LE &  CR/LE &  IN/LE &  IN/CE &  SP/HO &  SP/HO \\ \hline
\it{SA/V} &  CL/LE &  TO/LE &  CR/LE &  IN/LE &  SP/VE &  SA/CE &  LG/HO \\ \hline
\it{LO/V} &  CL/LE &  TO/LE &  CR/LE &  SP/LE &  SP/VE &  LG/VE &  LG/CE \\ \hline
\end{tabular}
   \caption{A part of $\mathbf{FAM_2}$ for estimating the 2-D fuzzy position descriptors (reduced due to space limitations, for complete 15x15 version see~\cite{Iwanowski2021}).}
   \label{tab:reasoning2}
\end{table}

\subsection{Scene description using FMP matrix}

Considerations in the previous section concern the case of relations between {\em two} objects. To get the description of the complete scene that consists of multiple objects, a matrix of one-to-one relations has been introduced, called the {\em fuzzy mutual position} ($FMP$) matrix. It consists of relative position descriptors for all combinations of bounding boxes. Let $\mathbf{S} = \left\{B^{(1)},B^{(2)}, ... ,B^{(n)} \right\}$ be the visual scene --  the set of $n$ bounding boxes of the image. The $FMP$ matrix is defined in $n \times n \times |D_2|$ as:

\begin{equation}
    FMP_{c,r,d}(\mathbf{S}) = \mu_d\left( B_{B^{(r)}}^{(c)} \right) \; \; 1 \leq c,r \leq n; d \in D_2,
    \label{eq:fmp}
\end{equation}

\noindent where $c$ and $r$ stands for objects' indexes: current and reference, respectively, and $d$ stands for the fuzzy 2-D descriptor\footnote{Due to the fact that, in general, fuzzy position descriptor is not symmetric i.e. $B_{B^{(j)}}^{(i)} \neq B_{B^{(i)}}^{(j)}$, $FMP$ is not symmetric neither.}. The $c$-th row consists of 2-D fuzzy descriptors of $c$-th object computed relative to all other objects considered as reference ones. The element of index $(c,r)$ contains a vector of descriptors of $B^{(c)}$ relative to $B^{(r)}$, and will be further referred to as $FMP_{c,r}(\mathbf{S})$.

\section{Similarity and symmetry detection}

The $FMP$ matrix describes the visual scene using the object's fuzzy descriptors of objects concerning other ones. One can say that two objects are in a similar position if their fuzzy 2-D descriptors are compliant. Let $B^{(i)}$ and $B^{(j)}$, be two object present on the first image ($i,j$ are their indexes).
They are also present on the second image (to be compared with the first one); let us call them $B^{\prime(i)}$ and $B^{\prime(j)}$. They may be located on both images in different places (their bounding boxes may have completely different image coordinates). We can then say that their compositions are similar if the fuzzy 2-D descriptors on both images (stored in the $FMP$ matrices) are close one to another i.e. $B_{B^{(j)}}^{(i)} \approx B_{B^{\prime(j)}}^{\prime(i)}$ and  $B_{B^{(i)}}^{(j)} \approx  B_{B^{\prime(i)}}^{\prime(j)}$. If we can draw the same conclusion from all other pairs $i,j$ (or, at least, for the vast majority of them), then the composition of objects on both images may be considered similar. The ratio of matched bounding boxes to all bounding boxes present within images may be proportional to the overall composition similarity. 

The above considerations may also be used to measure the symmetry of the image objects' composition. However, this time, the 2-D descriptor should be considered as composed two of two parts: locus-descriptor and orientation descriptor. The locus descriptor reflects an object's position relative to the reference one in terms of the fuzzy estimation of the relative distance to the reference object. It tells us whether the object is far, near, border-crossing, inside, etc. These properties are invariant to symmetry. There is, however, also the second part of the 2-D descriptor -- the orientation descriptor. This one, in turn, strongly depends on image symmetry because it describes the direction in which one may find the given object looking from the reference one.  In the symmetrical composition of object bounding boxes, these descriptors should be compared to their symmetrical counterparts. For example, we can say that the composition of some $B^{(i)}$ and $B^{(j)}$ is symmetric along x-axis, if $B^{(i)}$ is {\em close} to $B^{(j)}$ on both images but on the first it is located on the {\em left}, while on the second, on the {\em right}-hand side.  

The detailed description of the proposed approach to detecting similarity and symmetry is in the following subsections.

\subsection{Similarity detection}

Let $\mathbf{S}$ be the first scene (set of bounding boxes of the first image), where the total number of bounding boxes equals $n_1$. Let $\mathbf{S'}$ be the second scene -- the set of $n_2$ bounding boxes of the second image. Let, $id(B)$ be a function that returns the bounding box's ID-number. ID-numbers are unique numbers that identify the bounding boxes. We will use them to establish correspondences between objects. Each bounding box $B$ on the first image has its counterpart $B'$ on the second image iff $id(B) = id(B')$. Let assume that some number $n_0 > 0$ of bounding boxes from $\mathbf{S}$ and $\mathbf{S'}$ are matched, i.e., have the same ID number\footnote{In case $n_0 = 0$ we will not be able to estimate the similarity at all -- two images would be completely different, consisting of disjoint sets of objects. To estimate the similarity, at least two objects with fuzzy 2-D descriptors of their mutual relation must be present on both images with pairwise the same ID's.} The total number of different objects on both images equals thus to $n = n_1 + n_2 - n_0$. In the first step, objects present on both images are extracted to unified sets $\mathbf{C}$ and $\mathbf{C'}$ that, comparing to the original sets, are trimmed to the same length $n_0$ and ordered so that the bounding boxes of the same index have the same ID.

In order to estimate the total similarity of two images, in the beginning, the pairwise similarity between objects that are present on both images is evaluated by considering the appropriate elements of the mutual position matrices of both images. We thus reduce the problem of comparing two scenes to the process of investigating pairs of relations between sets of objects on two images. The $FMP$ matrix contains values of membership functions to all possible fuzzy 2-D descriptors fuzzy sets. For a given pair of objects, only a few (usually one or two) values are greater than 0. These descriptors indicate the possible relations between objects. Because membership functions' values indicate the strengths of those relations, the membership function values are thresholded, so the weakest relations are rejected. To compare a pair of relations between bounding boxes, one compares fuzzy descriptors of membership function values greater than a given threshold $t$. In order, in turn, to compare descriptors, each of two parts of the descriptors, locus and orientation, are investigated separately. For each part, one has to consider three cases. The first case refers to the equality of descriptor, i.e., the locus descriptor is equal to the same value on both images, e.g., "far". It is a clear case -- the distance is in the same range on both images, the locus-based similarity is complete. The second case refers to the situation when descriptor values are not equal but semantically close one to another, e.g., its value on the first image equals 'far' while on the second one -- to "near". In this case, a particular (lower) degree of similarity can also be assigned. Finally, there exists the third case. It refers to values that are semantically far from another (e.g. 'inside' vs. 'far'). In such a case, the similarity equals 0.

\begin{table}[]
    \centering \tiny
    \begin{tabular}{|c||c|c|c|c|c|c|c|c|c|}\hline 
            & FA & NE & CL & TO & CR & IN & LG & SP & SA \\ \hline \hline
         FA & 1  & vh & vl & 0  & 0  & 0  & 0  & 0  & 0  \\ \hline
         NE & vh & 1  & vh & vl & 0  & 0  & 0  & 0  & 0  \\ \hline
         CL & vl & vh & 1  & vh & vl & 0  & 0  & 0  & 0  \\ \hline        
         TO & 0  & vl & vh & 1  & vh & vl & 0  & 0  & 0  \\ \hline
         CR & 0  & 0  & vl & vh & 1  & vh & 0  & 0  & 0  \\ \hline
         IN & 0  & 0  & 0  & vl & vh & 1  & 0  & vl & vl \\ \hline
         LG & 0  & 0  & 0  & 0  & 0  & 0  & 1  & vl & vl \\ \hline
         SP & 0  & 0  & 0  & 0  & 0  & vl & vl & 1  & vl \\ \hline
         SA & 0  & 0  & 0  & 0  & 0  & vl & vl & vl & 1  \\ \hline
    \end{tabular}
    \caption{Matching matrix for locus descriptors $MM_{loc}$, intermediary values are equal to vh = 0.5 and vl = 0.25.}
   \label{tab:locmatch}
\end{table}

The same is true for orientation descriptors. Here one also should differentiate similarity degrees for cases of full (e.g. 'right' or both images), partial (e.g. 'right' on the first image, and 'right-above' on the second), and null (e.g. 'right' vs. 'above') accordance of descriptor values.

To properly and efficiently process the possible combinations of various descriptors' values, the {\em matching matrices $MM$} are used. There are two such matrices, one per each descriptor part (locus and orientation). Rows and columns of those matrices refer to descriptors' values to be matched. Their elements are real numbers between 0 and 1 which are weighting coefficients applied to evaluate a given combination's weight. Matching matrix for locus descriptor $MM_{loc}$ is presented in Table~\ref{tab:locmatch}. In the further considerations notation $MM_{loc}(d,d')$ is referring to the element of locus matching matrix for locus parts of descriptors $d$ and $d'$. Similarly, values in Table~\ref{tab:ormatch} are orientation matching values, further referred to as $MM_{or}(d,d')$.

The matching matrices are used as a kind of look-up-tables to compute the similarity between objects' relative positions on both images. The final similarity measure is computed as normalized {\em accumulated pairwise similarity}. The complete workflow of the method of its computing is presented as Algorithm 1. 

\begin{table}[]
    \centering \tiny
    \begin{tabular}{|r||c|c|c|c|c|c|c|c|c|c|c|}\hline 
            & LE    & LA    & AB    & RA    & RI    & RB    & BE    & LB    & CE    & HO   & VE    \\ \hline \hline
         1, LE & 1     & vh & 0     & 0     & 0     & 0     & 0     & h & 0    & vl & 0     \\ \hline
         2, LA & hi & 1     & vh & 0     & 0     & 0     & 0     & 0     & 0    & vl & vl \\ \hline
         3, AB & 0     & vh & 1     & vh & 0     & 0     & 0     & 0     & 0    & 0     & vl \\ \hline        
         4, RA & 0     & 0     & vh & 1     & vh & 0     & 0     & 0     & 0    & vl & vl \\ \hline
         5, RI & 0     & 0     & 0     & vh & 1     & vh & 0     & 0     & 0    & vl & 0     \\ \hline
         6, RB & 0     & 0     & 0     & 0     & vh & 1     & vh & 0     & 0    & vl & vl \\ \hline
         7, BE & 0     & 0     & 0     & 0     & 0     & vh & 1     & vh & 0    & 0     & vl \\ \hline
         8, LB & hi    & 0     & 0     & 0     & 0     & 0     & vl & 1     & 0    & vl & vl \\ \hline
         9, CE & 0     & 0     & 0     & 0     & 0     & 0     & 0     & 0     & 1    & vl     & vl     \\ \hline
         10, HO & vl   & vl    & 0     & vl    & vl    & vl    & 0     & vl     & vl    & 1     & 0  \\ \hline
         11, VE & 0    & vl    & vl    & vl    & 0     & vl    & vl    & vl     & vl    & 0     & 1  \\ \hline
    \end{tabular}
    \caption{Matching matrix for orientation descriptors $MM_{or}$, intermediary values are equal to vh = 0.5 and vl = 0.25.}
   \label{tab:ormatch}
\end{table}

\begin{tabular}{ll}
  & \\
  & {\bf Algorithm 1 -- computation of accumulated}  \\
  & {\bf pairwise similarity} \\ \hline
  & input data: $\mathbf{S}$,$\mathbf{S'}$  sets of bb's of two images\\
  & input parameter: $t$  threshold\\
  & output: $s_{acc}$ accumulated pairwise similarity\\ \hline
1 & compute $\mathbf{C}$, $\mathbf{C'}$, $n$ and $n_0$ based on $\mathbf{S}$ and $\mathbf{S'}$\\
2 & compute $FMP(\mathbf{C})$ and $FMP(\mathbf{C'})$\\
3 & $s_{acc} \leftarrow 0$\\
4 & for $i \leftarrow 1,...,n_0$:\\
5 & $\;\;$ for $j \leftarrow 1,...,n_0$:\\
6 &  $\;\;\;\;\; s \leftarrow 0$ \\
7 &  $\;\;\;\;$ for $d \leftarrow 1,...,|D_2|$: \\
8 &  $\;\;\;\;\;\;$ for $d' \leftarrow 1,...,|D_2|$: \\
9 &  $\;\;\;\;\;\;\;\;$ if $FMP_{i,j,d}(\mathbf{C}) > t  \wedge FMP_{i,j,d'}(\mathbf{C'}) > t$: \\
10 & $\;\;\;\;\;\;\;\;\;\;\; s_{loc} = MM_{loc}(d,d')$ \\
11 & $\;\;\;\;\;\;\;\;\;\;\; s_{or} = MM_{or}(d,d')$ \\
12 &  $\;\;\;\;\;\;\;\;\;\;\; s = \max(s,s_{loc} \cdot s_{or})$  \\
13 &  $\;\;\;\;\; s_{acc} \leftarrow s_{acc} + s$ \\ \hline \\
\end{tabular}

Two external loops of the algorithm (lines 4-5 up to 13) iterate all pairs of objects. For each pair of objects, their mutual position on both images is taken from their $FMP$ matrices to compute pairwise similarity measure -- stored in variable $s$ (initialized in line 6). These computations are performed in two internal loops (lines 7-8 up to 12). They iterate all combinations of 2-D descriptors of the current two objects to find all pairs of descriptors with membership function values greater than the given threshold $t$ (line 9). Each such finding means that there exists a pair of meaningful descriptors for which the pairwise similarity is computed as a product (line 12) of two values being the elements of two matching matrices. The first one reflects the pairwise similarity between loci (line 10, Table 3), while the second -- between orientations (line 11, Table 4). The final value of pairwise similarity is the highest value of all the above findings (line 12). The similarity values for all object pairs are summed up and finally returned as the accumulated pairwise similarity (stored in variable $s_{acc}$ initialized in line 3 and updated in line 13). 

The value of accumulated pairwise similarity, being the result of Algorithm 1, is used to finally compute the similarity measure of images (represented by sets of their proper bounding boxes) $\mathbf{S}$ and $\mathbf{S'}$ using the following normalization:

\begin{equation}
    sim(\mathbf{S},\mathbf{S'}) = \frac{s_{acc}}{n \cdot n_0}
\end{equation}

This expression represents the average value of pairwise similarity ($\frac{s_{acc}}{n_0^2}$) multiplied by a factor $\frac{n_0}{n}$ that is the ratio of matched objects. If all objects are matched, i.e., each object of one image has a counterpart on the second ($ n = n_0$), this factor equals 1. The final similarity measure is an average pairwise similarity. However, if some objects are present on only one image, the similarity factor is reduced accordingly. It follows the common-sense observation that the higher, among the total number of image objects, is the number of objects visible exclusively on one image, the lower is the images' similarity. 

\subsection{Reflectional symmetry detection}

Thanks to the separation of locus and orientation descriptors, the information on mutual relation directions can be analyzed independently on the locus information. In particular, one may observe very interesting results when analyzing the relation of the orientation descriptors and their relations to the composition of objects on both images being the subject of comparison. 

Let us imagine two similar visual scenes such that one is mirrored about another. The composition of objects in terms of their loci (how far are objects one from another, inside or outside others, etc.) is the same -- locus descriptors may have similar values. However, in orientation-descriptors, the relations on one image convert themselves to their counterparts on the other. The left-hand side moves to the right-hand side and vice-versa. Thus, the opposite ones replace all the orientation descriptors: {\em left} become {\em right} and reversely. 
    
The above properties can be encoded in the matching matrix $MM_{or}$. To get its version that measures symmetry, some of its rows must be interchanged (permuted) so that the best match is not with the same orientation descriptor on the second image but with its symmetric reflection. In case of classic mirroring, symmetry along x-axis, best match should be for the {\em left} $\leftrightarrow$ {\em right} correspondence. In the case of y-axis symmetry, the replacement should refer to  {\em above} $\leftrightarrow$ {\em below} correspondence and in the case of diagonal symmetry -- to both. 

Let $perm(MM, [a_1,a_2,...])$ be the operation that permutes rows of the $MM$ matrix so that the vector indicates their order in resulting matrix is the second argument, where $a_1,a_2,...$, are indexes of the original matrix: the first row of the permuted matrix is the $a_1$-th row of the original, the second row is the $a_2$-th, etc. The matching orientation matrices for three principal symmetries are the following (x-axis, y-axis, xy-axis, respectively):

\begin{equation}
\begin{split}
    MM_{or}^{x} = perm(MM_{or},[5,4,3,2,1,8,7,6,9,10,11]) \\
    MM_{or}^{y} = perm(MM_{or},[1,8,7,6,5,4,3,2,9,10,11) \\
    MM_{or}^{xy} = perm(MM_{or},[5,6,7,8,1,2,3,4,9,11,10]
\end{split}
\end{equation}

By replacing, in the Algorithm 1, the original $MM_{or}$ by one of the above variants, the measure obtained is proportional not to similarity of objects compositions, but to the given type of reflectional symmetry. 

\begin{figure}
 \centering
  \includegraphics[width=0.42\textwidth]{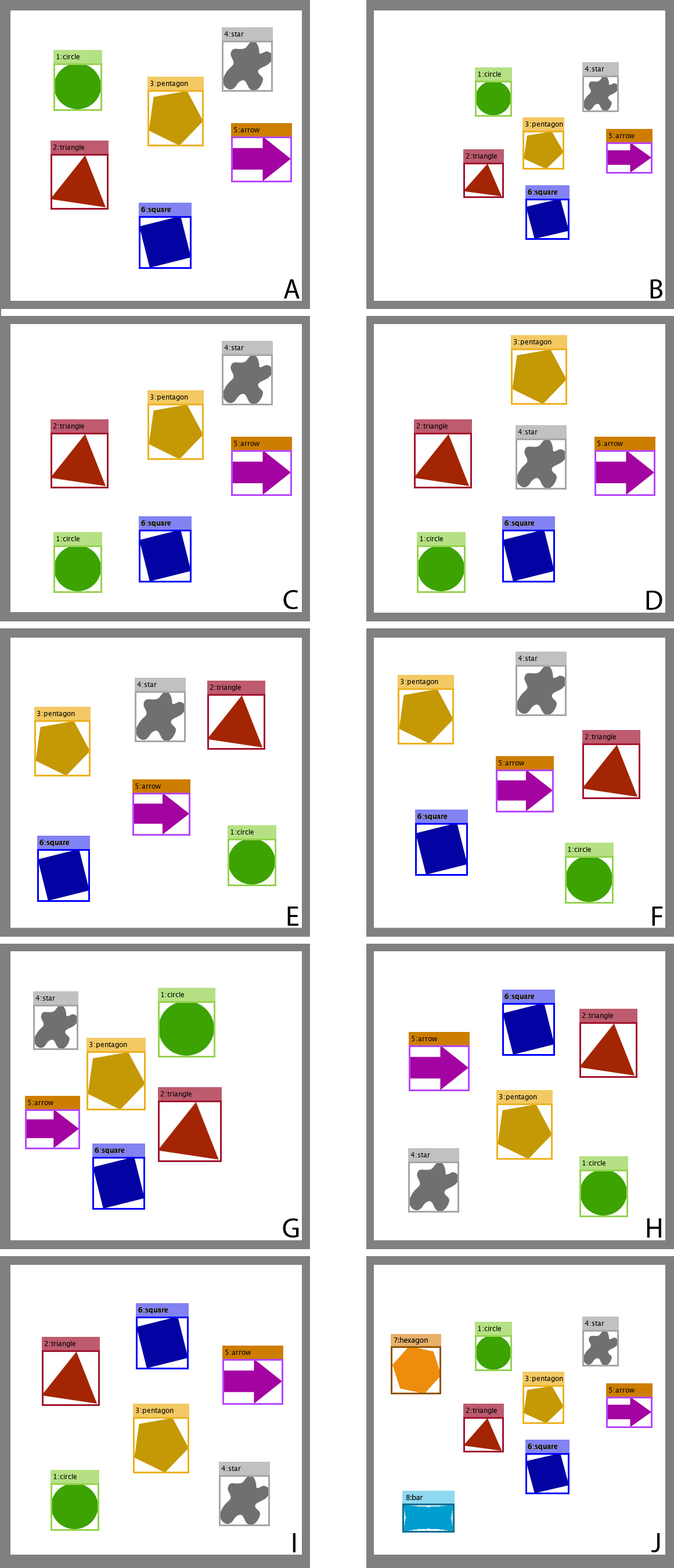} 
  \caption{Set of 9 artificial test images with 6 object classes: {\em circle}, {\em square}, {\em triangle}, {\em pentagon}, {\em arrow}, and {\em star}.}
  \label{fig:test1img}
\end{figure}

\begin{figure}
 \centering
  \includegraphics[width=0.45\textwidth]{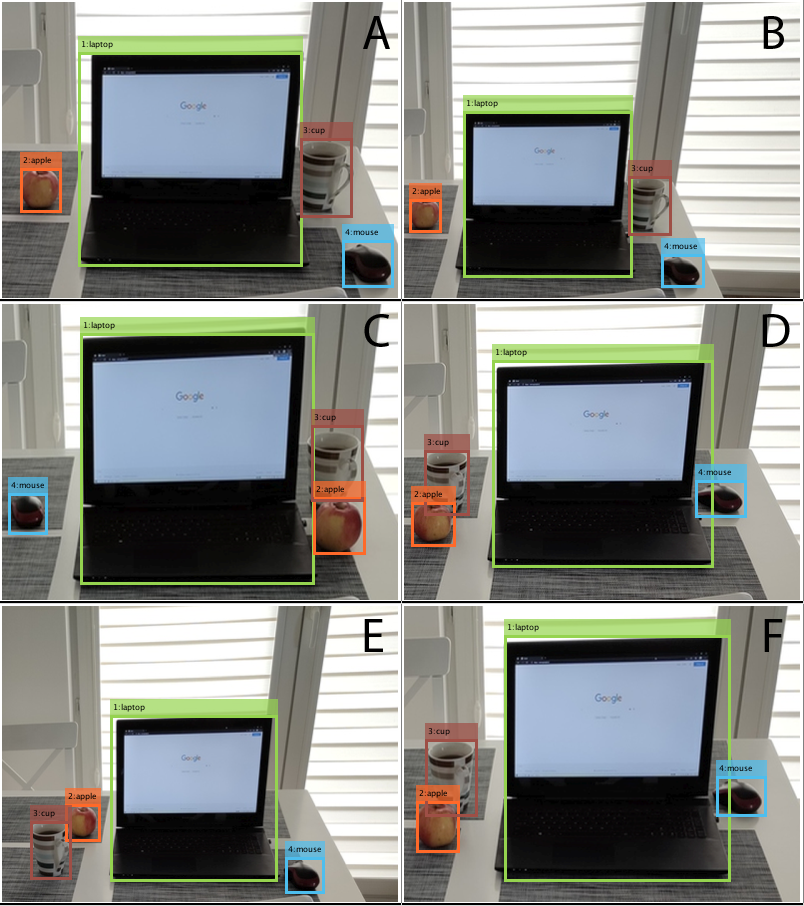} 
  \caption{Set of 6 real images processed using object detector (4 object classes detected on each image: {\em laptop}, {\em apple}, {\em mouse}, {\em cup}).}
  \label{fig:test2img}
\end{figure}

\section{Results}

The method has been tested on both artificially created images and images being the results of the object detection algorithm. The first set of scenes, shown in Fig.~\ref{fig:test1img} consist of 10 images consisting of 6 or 8 objects in different mutual positions. The differences are of various kind:  modifications of absolute position without changing their mutual positions considerably (see image A vs. B or E vs. F), two object replacement (see A vs. C, C vs. D), more replacements (e.g., B vs. D), x-symmetry (e.g., A vs. G), y-symmetry (A vs. I), different number of objects (B vs. J). The similarity and symmetry measures for all pairs of images are given in Tables~\ref{tab:similarity},~\ref{tab:symmetry_x},~\ref{tab:symmetry_y},~\ref{tab:symmetry_xy}.

\begin{table}[]
    \centering \tiny
    \begin{tabular}{|c|c|c|c|c|c|c|c|c|c|c|}\hline
   	A	&	B	&	C	&	D	&	E	&	F	&	G	&	H	&	I	&  J	&	\\	\hline
	1.00	&	{\bf 0.97}	&	{\bf 0.76}	&	0.53	&	0.44	&	0.44	&	0.33	&	0.17	&	0.39	&	0.72	&  A	\\	
	-	&	1.00	&	{\bf 0.76}	&	0.50	&	0.44	&	0.44	&	0.33	&	0.17	&	0.39	& 0.75 &	B	\\	
	-	&	-	&	1.00	&	0.71	&	0.47	&	0.53	&	0.28	&	0.19	&	0.51	& 0.57 &	C	\\	
	-	&	-	&	-	&	1.00	&	0.36	&	0.42	&	0.21	&	0.22	&	0.47	& 0.38	& D	\\	
	-	&	-	&	-	&	-	&	1.00	&	0.89	&	0.41	&	0.39	&	0.28	&	0.33 & E	\\	
	-	&	-	&	-	&	-	&	-	&	1.00	&	0.44	&	0.33	&	0.26	&	0.33 & F	\\	
	-	&	-	&	-	&	-	&	-	&	-	&	1.00	&	0.39	&	0.17	& 0.25 & G	\\	
	-	&	-	&	-	&	-	&	-	&	-	&	-	&	1.00	&	0.33	& 0.13 &H	\\	
	-	&	-	&	-	&	-	&	-	&	-	&	-	&	-	&	1.00	&	0.29 &  I	\\	
	-   &	-	&	-	&	-	&	-	&	-	&	-	&	-	&	-	    &	1.00	& J	\\	\hline
    \end{tabular}
    \caption{Similarity measures of pairs of images from the test set 1 shown in Fig.~\ref{fig:test1img}.}
   \label{tab:similarity}
\end{table}

\begin{table}[]
    \centering  \tiny
    \begin{tabular}{|c|c|c|c|c|c|c|}\hline
A	 &	B	&	C	&	D	&	E	&	F	&	 \\ \hline
1.00 & {\bf 1.00} & 0.38	& 0.49	& 0.62  & 0.49 &  A \\
 -   &  1.00      & 0.38	& 0.50	& 0.56	& 0.50 &  B \\
 -	 &	-      	  & 1.00	& 0.31	& 0.25 	& 0.34 &  C \\
 -	 &	-	      &	-	& 1.00	& 0.73	& {\bf 1.00} &  D \\
 -	 &	-	      &	-	&	-	& 1.00 	& 0.73 &  E \\
 -	 &	-	      &	-	&	-	&	-	& 1.00 &  F \\ \hline
    \end{tabular}
    \caption{Measures of similarity of image pairs (test set 2 shown in Fig.~\ref{fig:test2img})}
   \label{tab:similarity2}
\end{table}

The second example scenes are real images presenting the same vision scene consisting of four objects: {\em laptop}, {\em apple}, {\em mouse}, {\em cup} automatically annotated using object detection method (see Fig.~\ref{fig:test2img}). Here also various configurations of objects are present. We can observe a variation of scale (see image A vs. B), the various position of objects (A and B vs. C, D, E), as well as mirror symmetry (x-axis, see C vs. F). The only symmetry observed here is mirroring, so the matrix presenting the x-axis symmetry is presented (see Table~\ref{tab:symmetry2}) along with the similarity measures array (see Table~\ref{tab:similarity2}).

\begin{table}[]
    \centering \tiny
    \begin{tabular}{|c|c|c|c|c|c|c|c|c|c|c|}\hline
A	&	B	&	C	&	D	&	E	&	F	&	G	&	H	&	I	&	J  &	\\	\hline
0.33	&	0.33	&	0.28	&	0.21	&	0.41	&	0.42	&	{\bf 0.85}	&	0.39 &0.17	&	0.25	&	A	\\	
-	&	0.33	&	0.28	&	0.21	&	0.41	&	0.42	&	{\bf 0.89}	&	0.39 &0.17	&	0.25	&	B	\\	
-	&	-	&	0.33	&	0.27	&	0.55	&	0.58	&	0.65	&	0.50	&	0.21	&	0.21 & C	\\	
-	&	-	&	-	&	0.39	&	0.49	&	0.51	&	0.42	&	0.46	&	0.24	& 0.16 & D	\\	
-	&	-	&	-	&	-	&	0.33	&	0.33	&	0.40	&	0.28	&	0.39	&	0.31 & E	\\	
-	&	-	&	-	&	-	&	-	&	0.39	&	0.43	&	0.25	&	0.35	& 0.21 &	F	\\	
-	&	-	&	-	&	-	&	-	&	-	&	0.33	&	0.17	&	0.33	&	0.67 & G	\\	
-	&	-	&	-	&	-	&	-	&	-	&	-	&	0.33	&	{\bf 1.00}	&	0.29 & H	\\	
-	&	-	&	-	&	-	&	-	&	-	&	-	&	-	&	0.33	&  0.13 & I	\\	
-	&	-	&	-	&	-	&	-	&	-	&	-	&	-	&	-	&  0.25 &	J	\\	\hline
    \end{tabular}
    \caption{Measures of symmetry (x-axis) of image pairs (test set 1 shown in Fig.~\ref{fig:test1img}).}
   \label{tab:symmetry_x}
\end{table}

\begin{table}[]
    \centering \tiny
    \begin{tabular}{|c|c|c|c|c|c|c|c|c|c|c|}\hline
A	&	B	&	C	&	D	&	E	&	F	&	G	& H &	I	&	J	&	\\ \hline
0.39	&	0.38	&	0.48	&	0.43	&	0.24	&	0.27	&	0.17	&	0.29	&	{\bf 0.88}	&	0.27 &  A \\
-	&	0.39	&	0.51	&	0.48	&	0.25	&	0.28	&	0.17	&	0.29	&	{\bf 0.89}	&	0.29 & B\\
-	&	-	&	0.39	&	0.36	&	0.21	&	0.21	&	0.22	&	0.26	&	0.65	& 0.39 & C\\
-	&	-	&	-	&	0.39	&	0.17	&	0.17	&	0.24	&	0.21	&	0.49	&	0.36 & D\\
-	&	-	&	-	&	-	&	0.39	&	0.39	&	0.38	&	0.37	&	0.44	& 0.19 &	E\\
-	&	-	&	-	&	-	&	-	&	0.39	&	0.40	&	0.38	&	0.44	& 0.21 & 	F\\
-	&	-	&	-	&	-	&	-	&	-	&	0.44	&	0.86	&	0.32	&	0.13 & G\\
-	&	-	&	-	&	-	&	-	&	-	&	-	&	0.39	&	0.17	&	0.22 & H\\
-	&	-	&	-	&	-	&	-	&	-	&	-	&	-	&	0.39	&	0.67 & I\\	
-	&	-	&	-	&	-	&	-	&	-	&	-	&	-	&	-	&	0.29 & J\\	\hline
    \end{tabular}
    \caption{Measures of symmetry (y-axis) of image pairs (test set 1 shown in Fig.~\ref{fig:test1img}).}
   \label{tab:symmetry_y}
\end{table}

\begin{table}[]
    \centering \tiny
    \begin{tabular}{|c|c|c|c|c|c|c|c|c|c|c|}\hline
A	&	B	&	C	&	D	&	E	&	F	&	G	& H & 	I	&	J	&	\\ \hline
0.17	&	0.17	&	0.22	&	0.24	&	0.33	&	0.33	&	0.35	&	{\bf 0.85}	&	0.31 & 0.13 	&	A\\
-	&	0.17	&	0.22	&	0.25	&	0.35	&	0.35	&	0.39	&	{\bf 0.89}	&	0.31	& 0.13 &	B\\
-	&	-	&	0.17	&	0.18	&	0.25	&	0.25	&	0.44	&	0.64	&	0.26 &  0.17	&	C\\
-	&	-	&	-	&	0.17	&	0.26	&	0.26	&	0.42	&	0.49	&	0.21	&	0.19 & D\\
-	&	-	&	-	&	-	&	0.17	&	0.17	&	0.21	&	0.44	&	0.37	& 0.26 & E\\
-	&	-	&	-	&	-	&	-	&	0.17	&	0.23	&	0.44	&	0.38	&	0.27 & F\\
-	&	-	&	-	&	-	&	-	&	-	&	0.17	&	0.31	&	{\bf 0.80}	&	0.29 & G\\
-	&	-	&	-	&	-	&	-	&	-	&	-	&	0.17	&	0.39	&	0.67 & H\\
-	&	-	&	-	&	-	&	-	&	-	&	-	&	-	&	0.17	&	0.23 & I	\\	\
-	&	-	&	-	&	-	&	-	&	-	&	-	&	-	&	-	&	0.13 & J	\\	\hline
    \end{tabular}
    \caption{Measures of symmetry (x- and y-axis) of image pairs (test set 1 shown in Fig.~\ref{fig:test1img}).}
   \label{tab:symmetry_xy}
\end{table}

\begin{table}[]
    \centering \tiny
    \begin{tabular}{|c|c|c|c|c|c|c|}\hline
 A    & 	B	&	C	    &	D	&	E	&	F	&	\\ \hline
 0.38 &  0.38 	&	0.55	& 0.34	& 0.30  & 0.34	&  A \\
 -	  &	 0.38	&	0.52	& 0.34	& 0.30  & 0.34	&  B \\
 -	  &	 -	    &	0.38	& {\bf 0.86} & 0.67  & 0.83  &  C \\
 -	  &	-    	&	-	    &	0.25	&	0.25	& 0.28  &  D \\
 -	  &	-	    &	-	    &	-	   &	0.25	&  0.25    &  E \\
  -	  &	-	    &	-	    &	-	   &	-	&  0.31 &  F \\ \hline
    \end{tabular}
    \caption{Measures of symmetry (x-axis) of image pairs (test set 2 shown in Fig.~\ref{fig:test2img})}
   \label{tab:symmetry2}
\end{table}

The discussion of the results is covered in following items pointing out properties of the proposed approach:

\begin{itemize}
    \item {\bf Translation and scale invariance} Thanks to the very first step of processing -- conversion of image coordinates to relative ones, the method is scale and translation invariant. Both those transformation does not change the values of 2-D descriptors as long as the composition of objects within their group is the same. Examples confirming this property are the following\footnote{To simplify explanation, in the references we will use shortcuts. For example, ,,1A vs.1B Tab.~\ref{tab:similarity} $sim=0.97$'' means that image from the test scenes 1 (shown in Fig.3) ,,A'' should be compared with an image shown on the position ,,B'', the relevant measures are in Table~\ref{tab:similarity}, the similarity measure in this case is equal to 0.97. Images from the second set of scenes (shown in Fig.4) will be referred to as 2A, 2B, etc.}: 1A vs.1B Tab.~\ref{tab:similarity} $sim=0.97$; 2A vs.2B Tab.~\ref{tab:similarity2} $sim=1.00$; 2D vs.2F Tab.~\ref{tab:similarity2} $sim=1.00$. 
    \item {\bf Proportionality to the scale of changes}. The measure decreases when the number of objects that changed their positions grows. The lower is thus the visual similarity (judged by a human observer), the lower is also the computed measure. Examples: 1A vs.1C (one replacement of objects) Tab.~\ref{tab:similarity} $sim=0.76$; 1A vs.1D (two replacements) Tab.~\ref{tab:similarity} $sim=0.53$; 1A vs.1H (completely different composition) Tab.~\ref{tab:similarity} $sim=0.17$; 2A vs. 2E (one shift) Tab.~\ref{tab:similarity2} $sim=0.62$; 2A vs.2C (three shifts) Tab.~\ref{tab:similarity2} $sim=0.38$.
    \item {\bf Penalizing non-matched object} Thanks to the last step of processing, the normalization of accumulated pairwise similarity, the presence of objects on only one image reduces the similarity measure. Example: 1A vs.1J (two objects added, the remainder in the same composition) Tab.~\ref{tab:similarity} $sim=0.72$, to compare with 1A vs.1B.
    \item {\bf Ability to detect symmetries} When choosing the appropriate matching matrices, the method is able to detect three reflectional symmetries. Examples, x-axis: 1A vs.1G Tab.~\ref{tab:symmetry_x} $sim=0.85$; 1I vs.1H Tab.~\ref{tab:symmetry_x} $sim=1$; 2C vs.2D Tab.~\ref{tab:symmetry2} $sim=0.86$; y-axis: 1A vs.1I Tab.~\ref{tab:symmetry_y} $sim=0.88$; 1B vs.1I Tab.~\ref{tab:symmetry_y} $sim=0.89$; xy-axis: 1A vs.1H Tab.~\ref{tab:symmetry_xy} sim=0.85; 1B vs.1H Tab.~\ref{tab:symmetry_xy} $sim=0.89$; 1I vs.1G Tab.~\ref{tab:symmetry_xy} $sim=0.80$.
    \item {\bf Axis of symmetry position's invariant symmetry measuring} Thanks to the relative way of describing the position of objects, the method is able to detect symmetries no matter where the symmetry axis is located. Example for x-axis: 1A vs.1G Tab.~\ref{tab:symmetry_x} $sim=0.85$; 1B vs.1G Tab.~\ref{tab:symmetry_x} $sim=0.89$.
\end{itemize}

\begin{table}[]
    \centering  \tiny
    \begin{tabular}{|c|c|c|c|c|c|c|c|c|c|c|}\hline
method	&	B,A	&	C,A	&	D,A	&	E,A	&	F,A	&	G,A	&	H,A	&	I,A	&	J,A	\\	\hline
ours	&	0.97&	0.76&	0.53&	0.44&	0.44&	0.33&	0.17&	0.39&	0.72	\\	
MSE	    &	0.08&	0.04&	0.07&	0.12&	0.1	&	0.11&	0.11&	0.12&	0.08	\\	
PSNR	&	10.78	& 14.01	& 11.87	& 9.16	& 9.88	& 9.42	& 9.41	& 9.38	& 10.78	\\	
SSIM	&	0.73 & 0.88	& 0.78	& 0.62	& 0.66	& 0.65	& 0.63	& 0.63	& 0.73	\\	
M-SSIM	&	0.62 & 0.85	& 0.71	& 0.5	& 0.54	& 0.54	& 0.52	& 0.52	& 0.62	\\	\hline
    \end{tabular}
    \caption{Results of the proposed method ("ours") and standard similarity measures, similarity measures of image A with all other (test set 1 shown in Fig.~\ref{fig:test1img})}
   \label{tab:measures1}
\end{table}

\begin{table}[]
    \centering  \tiny
    \begin{tabular}{|c|c|c|c|c|c|}\hline
	method	& B,A & C,A & D,A & E,A & F,A \\	\hline
    ours	& 1	   & 0.38 & 0.49 & 0.62 & 0.49	\\	
    MSE	    & 0.15 & 0.07 &	0.04 & 0.13	& 0.07	\\	
    PSNR	& 8.36 & 11.77&	14.05& 8.88	& 11.56	\\	
    SSIM	& 0.24 & 0.48 &	0.51 & 0.3	& 0.45	\\	
    M-SSIM	& 0.16 & 0.47 &	0.56 & 0.2	& 0.39	\\	\hline
    \end{tabular}
    \caption{Results of the proposed method ("ours") and standard similarity measures, similarity measures of image A with all other (test set 2 shown in Fig.~\ref{fig:test2img})}
   \label{tab:measures2}
\end{table}

The proposed approach was also experimentally compared to classic similarity measures: mean square error (MSE), peak signal-to-noise ratio (PSNR), structural similarity index (SSIM)~\cite{Zhang1993}, and its multiscale variant (M-SSIM)~\cite{Wang2003}. Results are shown in Figs.~\ref{tab:measures1} and~\ref{tab:measures2}. They show the values of measures computed with scene 1 and scene 2. In each case, the first scene (A) is compared with all others (B, C,...). Contrary to the proposed approach (first row), values of the remainder of measures do not follow the variations of the complexity of the composition of objects (bounding boxes). When looking for the visual similarity sensed by a human, one may introduce the following order of scenes starting from the most similar to scene A: for the set 1: B (scale), C (one replacement), J (two objects added), D (two replacements), E, F (four objects in the same mutual position), I (two groups of objects in similar mutual position), G, H (completely different composition) and for the scene 2: B (scale), E (one object shifted to the other side), D, F (in both one shifted to the other side, one shifted closed to the central one), C (two objects shifted). The proposed measure follows this order, while none of the classic ones does. What is more, considering the above ordering based on human perception, the ordering implied by classic measures seems to besomehow random. 

\section{Conclusions}

The paper describes a novel method for measuring the similarity and symmetry of the annotated images' content consisting of objects defined by their bounding boxes. It allows comparing sets of bounding boxes to estimate the degree of similarity of their underlying images. It is based on the fuzzy approach that uses the fuzzy mutual positions matrix to describe spatial composition and relations between bounding boxes within a single image. In the paper, a method and algorithm for comparing such matrices computed for two images are proposed. It allows measuring the similarity of two images and outputs the single scalar value describing the degree of similarity. Modification of matching matrices used to establish correspondence between 2-D scene descriptors allows for applying the method to estimate the reflectional symmetries in the composition of the objects of the visual scenes. 

The method has several valuable properties. It is translation and scale-invariant. The resulting measure is proportional to the degree of differences between images. It can detect symmetries no matter where the symmetry axis is located. It also has few parameters that allow its adjusting to particular purposes. 

Moreover, it may be easily extended to get extra functionalities like, e.g., the ability to find the symmetric subsets of image objects. Last but not least, the method is fast and ready to use in real-time applications. It is due to the fact that it consists of simple operations: arithmetic operations, min/max, performed in nested loops but iterated with a relatively small number of times.  It may be used in content-based image retrieval, semantic image analysis, and other computer vision fields.


\printbibliography

\end{document}